\documentclass[10pt,twocolumn,letterpaper]{article}
\usepackage[T1]{fontenc}
\usepackage[pagenumbers]{cvpr}

\newcommand{\topflowmodel}{TOPFlow}
\newcommand{\densemeshset}{DenseMesh}

\newcommand{\addedvertexpercent}{0.034\%}

\usepackage{colortbl}
\definecolor{papergray}{HTML}{F0F2F5}
\usepackage{placeins}

\definecolor{cvprblue}{rgb}{0.21,0.49,0.74}
\usepackage[pagebackref,breaklinks,colorlinks,allcolors=cvprblue]{hyperref}
\hypersetup{
    pdftitle={TOPFlow: Tree-Structured Occupancy Posterior Flow for Mesh Vertex Generation},
    pdfauthor={Kaiyi Zhang, Dehao Hao, Haohan Weng, Xianghui Yang, Zhuo Chen, Long Quan},
    pdfsubject={Mesh vertex generation},
    pdfkeywords={3D generation, mesh generation, octree, flow matching}
}

\title{TOPFlow: Tree-Structured Occupancy Posterior Flow for Mesh Vertex Generation}
\author{%
    Kaiyi Zhang\textsuperscript{1,*}\quad
    Dehao Hao\textsuperscript{1,*}\quad
    Haohan Weng\textsuperscript{2}\quad
    Xianghui Yang\textsuperscript{2}\quad
    Zhuo Chen\textsuperscript{2}\quad
    Long Quan\textsuperscript{1,\textdagger}\\
    \textsuperscript{1}HKUST\quad
    \textsuperscript{2}Tencent Hunyuan\\
    {\ttfamily\small\{kzhangca,dhaoab,quan\}@cse.ust.hk}\\
    {\small\textsuperscript{*}Equal contribution\quad
    \textsuperscript{\textdagger}Corresponding author}
}
\date{}

\makeatletter
\g@addto@macro\@maketitle{\begin{center}
\vspace{-16pt}
\includegraphics[width=\textwidth]{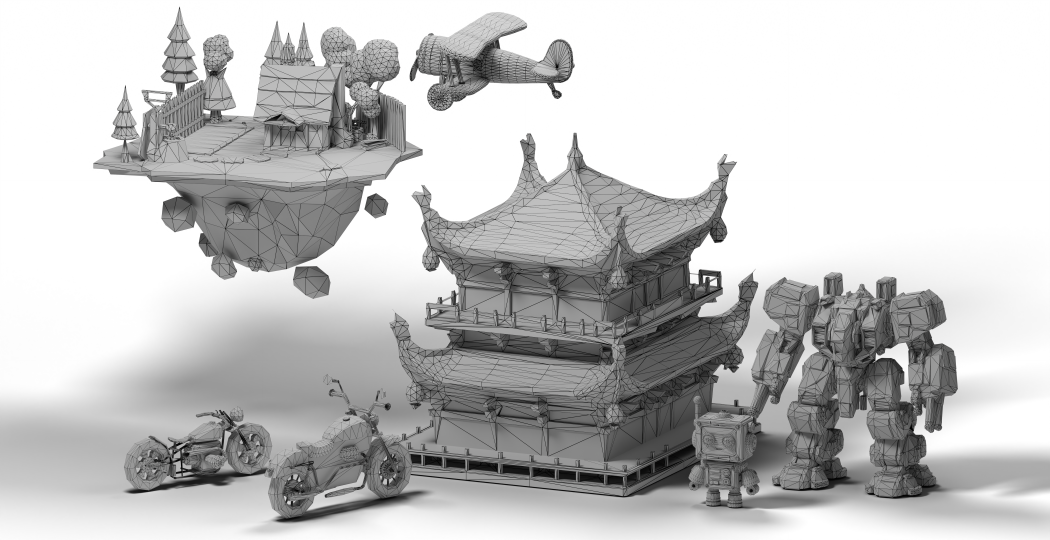}
\captionof{figure}{\textbf{Meshes generated by TOPFlow.}
Our results showcase fine geometric details and coherent structures across diverse objects.}
\label{fig:teaser}
\end{center}
}
\makeatother

\begin{document}
\maketitle
\begin{abstract}
Accurate mesh vertex generation requires capturing spatial dependencies to preserve geometric detail and coherent local structure.
We propose TOPFlow, a Tree-Structured Occupancy Posterior Flow framework for low-poly mesh vertex generation.
TOPFlow represents vertex sets as sparse octree occupancy and learns a joint posterior over valid child-cell occupancy patterns.
A shared spatial conditional tree uses global shape context and the complete preceding decision history to model dependencies across spatial subregions.
Maximizing complete-pattern likelihood provides discrete supervision, while exact computation of the posterior mean drives coarse-to-fine continuous generation.
The same posterior supports independent temperature control over occupied-child count and spatial layout at coarse and fine scales.
Experiments on Objaverse, Toys4K, and \densemeshset{} demonstrate improved vertex layouts and final-mesh geometric fidelity.
Against an octree flow baseline, TOPFlow reduces density-aware Chamfer distance by 14.2--17.4\% and final-mesh Hausdorff distance by 26.0--35.7\% on Objaverse and Toys4K, with only a \addedvertexpercent{} increase in vertex-generator parameters.
It achieves the best final-mesh Chamfer distance and normal consistency among the evaluated autoregressive and diffusion-based methods across all three datasets.
\end{abstract}

\section{Introduction}
\label{sec:intro}

Low-poly mesh generation seeks to represent an object's geometry through a compact set of vertices and their connectivity.
Recent approaches model meshes through autoregressive sequences~\cite{meshgpt2023,meshanything2024,bpt2024,meshripple2025} or generate meshes with diffusion and flow matching~\cite{polydiff2023,meshflowmeta2026}.
The spatial arrangement of vertices determines where geometric detail can be represented and provides the foundation for subsequent connectivity.
Capturing spatial dependencies is therefore central to distributing a compact vertex set across both overall shape and local detail.

Spatial hierarchies provide a natural way to organize geometry across scales.
OctFusion combines octree latent representations with multiscale diffusion~\cite{xiong2025octfusion}, while OctGPT uses serialized octrees and multiscale autoregression for shape generation~\cite{wei2025octgpt}.
For mesh generation, spatial refinement operates through octree vertex occupancy in Nexus~\cite{nexus2026} or vertex split-and-rewire operations in MeshOctave~\cite{meshoctave2026}.
We view an individual octree refinement as a joint decision about how vertices are distributed within a parent cell.
Its eight child occupancies jointly determine the occupied-child count and spatial layout, establishing the support available to subsequent refinement.
Choices in one subregion must therefore be modeled in relation to those in other subregions.
This motivates a spatially structured model of joint occupancy within each refinement level.

To model these coupled choices, we propose \topflowmodel{}, \emph{Tree-Structured Occupancy Posterior Flow}, for low-poly mesh vertex generation conditioned on point clouds.
A shared spatial conditional tree introduces a hierarchy of decisions within each octree refinement level.
Each cut predicts whether the left subregion, the right subregion, or both are nonempty, conditioned on the parent's global backbone context and the complete history of preceding local decisions.
This history includes sibling regions and other branches, allowing each cut to depend on spatial choices beyond its own ancestor path.
The conditional probabilities define a normalized joint posterior over the 255 nonempty child-cell patterns.
This finite local support enables complete evaluation in a small shared head while reusing the backbone context.

This explicit posterior provides a common basis for supervision, continuous generation, and geometric control.
Training maximizes complete-pattern likelihood, directly supervising configuration probabilities; optimal squared-error regression recovers their conditional mean~\cite{proper2007}, which alone does not uniquely determine the joint distribution.
The discrete occupancy outcomes remain compatible with a continuous generative state~\cite{analogbits2022}.
For a linear noise-to-occupancy path, the posterior mean determines the marginal flow velocity~\cite{flowmatching2022}, and TOPFlow computes this mean exactly for its learned distribution.
The same posterior further factors into occupied-child count, coarse spatial allocation, and fine layout, providing independent temperature controls.
These controls reshape the posterior mean and hence the continuous update, connecting uncertainty in local configurations to the generated vertex layout.

Experiments on Toys4K, Objaverse, and \densemeshset{} evaluate vertex layouts and final-mesh geometric fidelity.
With 40K conditioning points, \topflowmodel{} reduces density-aware Chamfer distance by 14.2--17.4\% and final-mesh Hausdorff distance by 26.0--35.7\% on Objaverse and Toys4K relative to our Nexus reimplementation, with only a \addedvertexpercent{} increase in vertex-generator parameters.
It achieves the best final-mesh Chamfer distance and normal consistency among the evaluated autoregressive and diffusion-based methods across all three datasets.
Our contributions are:
\begin{itemize}
    \item A shared spatial conditional tree that organizes occupancy decisions within each parent cell, using global shape context and the complete history of preceding spatial choices.
    \item A joint occupancy posterior trained through complete-pattern likelihood, whose exactly computed mean connects discrete supervision with continuous flow for vertex generation.
    \item Separate temperatures from a shared posterior control uncertainty in occupied-child count, coarse spatial allocation, and fine layout during generation.
\end{itemize}

\section{Related Work}
\label{sec:related_work}

\paragraph{Continuous and discrete diffusion models.}
Continuous diffusion models generate data by reversing a gradual noising process.
DDPM~\cite{ddpm2020} provides a denoising formulation, while score-based SDEs~\cite{scoresde2020} unify continuous-time diffusion and probability-flow ODEs.
Flow Matching~\cite{flowmatching2022} and Rectified Flow~\cite{rectifiedflow2022} learn vector fields that transport noise to data along prescribed probability paths.
For discrete states, D3PM~\cite{d3pm2021} uses categorical transitions, while continuous-time discrete denoising~\cite{discretedenoising2022} formulates generation through Markov jump processes.
Discrete Flow Matching~\cite{discreteflowmatching2024} learns dynamics along general discrete probability paths.
Analog Bits~\cite{analogbits2022} instead represents discrete targets as bits, generates them with continuous diffusion, and thresholds the resulting values for decoding.

\paragraph{Autoregressive mesh generation.}
Autoregressive methods serialize mesh geometry and topology into token sequences~\cite{polygen2020,meshxl2024,meshtron2024}.
MeshGPT~\cite{meshgpt2023} learns a quantized local mesh vocabulary on ShapeNet~\cite{shapenet2015}, while MeshAnything and its successor extend this approach to shape-conditioned synthesis~\cite{meshanything2024,meshanythingv22024}.
Subsequent tokenizers reduce coordinate redundancy and exploit connected surface structure~\cite{edgerunner2024,bpt2024,nautilus2025,freemesh2025,xu2026sato,song2026silksong}.
TreeMeshGPT~\cite{treemeshgpt2025} and MeshRipple~\cite{meshripple2025} further organize geometric context through face-adjacency trees and active surface frontiers, respectively.
Other extensions explore controllable refinement, improved training and decoding, and voxel-guided generation~\cite{zhang2025vertexregen,lei2025armesh,deepmesh2025,fastmesh2025,meshweaver2026}.
These methods emphasize compact representations and geometric context.

\paragraph{Diffusion-based mesh generation.}
Diffusion and flow models offer parallel generation through triangle soups~\cite{polydiff2023,meshflowequivariant2026}, continuous connectivity or topology fields~\cite{spacemesh2024,triflow2026}, and learned representations of mesh geometry and topology~\cite{meshcraft2025,polyflow2026,lato2026}.
MeshFlow~\cite{meshflowmeta2026} jointly represents vertex positions and connectivity in MeshVAE latents, while LATO.2~\cite{lato22026} factorizes their flows around a shared voxel scaffold.
Spatial hierarchies support multiscale shape generation through octree diffusion and autoregression~\cite{xiong2025octfusion,wei2025octgpt}.
Nexus~\cite{nexus2026} applies continuous flow to octree vertex occupancy before predicting spacetime connectivity.
Other hierarchical methods populate voxel scaffolds with vertex latents or refine geometry and connectivity through split-and-rewire operations~\cite{meshyt22026,meshoctave2026}.
Within octree refinement, TOPFlow models the joint child-occupancy posterior.

\section{Method}
\label{sec:method}

\begin{figure*}[t]
    \centering
    \includegraphics[width=\textwidth]{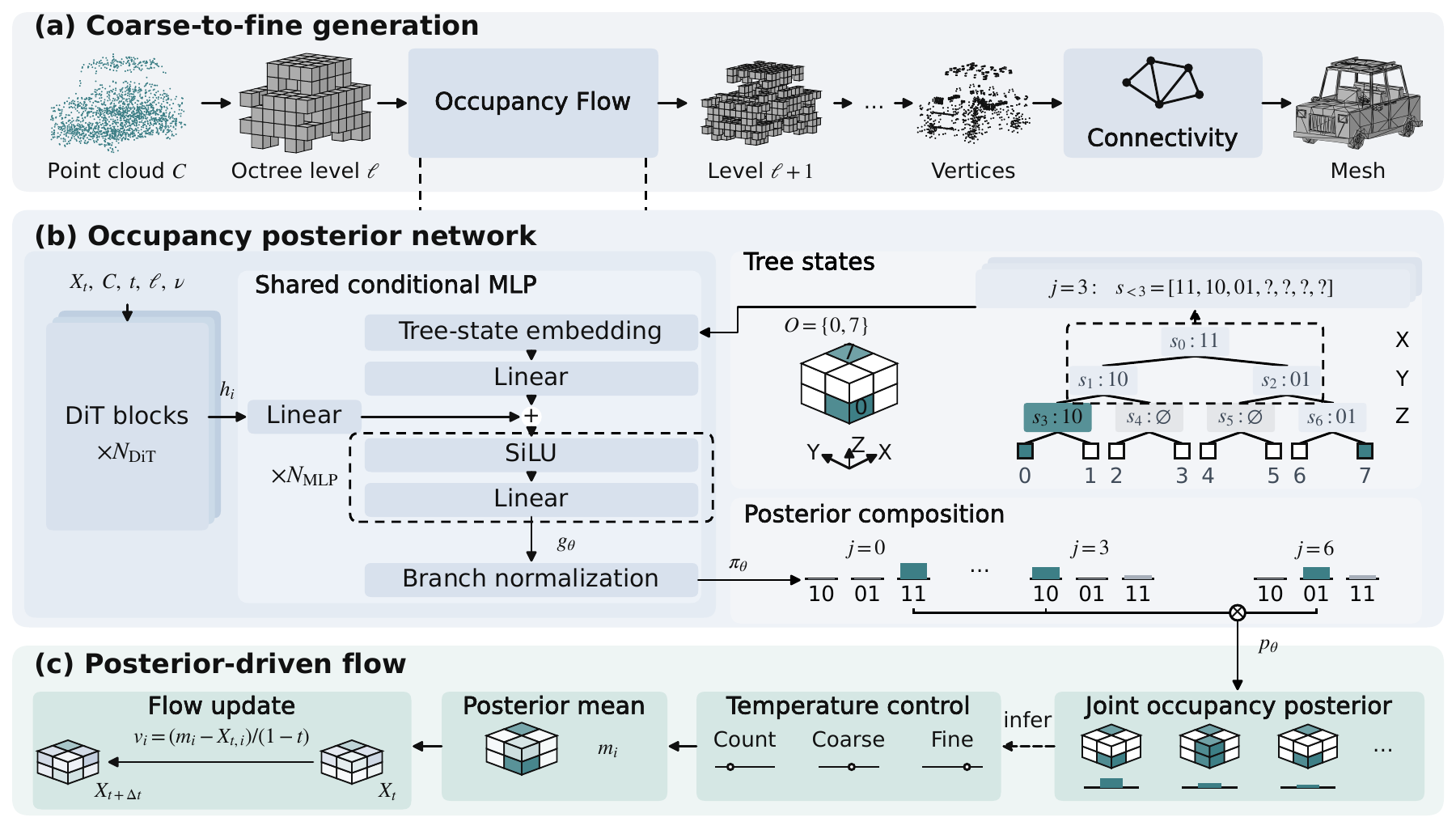}
    \caption{\textbf{TOPFlow overview.}
    (a) Coarse-to-fine octree refinement generates vertices, followed by connectivity prediction.
    (b) A shared DiT supplies global context to a conditional MLP evaluated across candidate tree states. Branch normalization yields three-way probabilities; posterior composition multiplies selected factors along valid decision sequences to form the joint occupancy posterior.
    (c) The temperature-controlled posterior yields an exactly computed mean that drives continuous flow updates.}
    \label{fig:method}
\end{figure*}

TOPFlow generates low-poly mesh vertices conditioned on a point cloud by progressively refining occupied octree cells.
Within each refinement level, a shared spatial conditional tree models the joint occupancy of every parent's eight children using global context and complete local decision histories.
Complete-pattern likelihood trains this distribution, whose exactly computed mean drives continuous generation; separate temperatures control count and multiscale layout.
\cref{fig:method} summarizes the architecture.

\subsection{Tree-Structured Occupancy Posterior}
\label{sec:tree_posterior}

\paragraph{Occupancy representation.}
Let $C$ be the conditioning point cloud with surface normals and $\mathcal F_\ell$ the occupied parent cells at octree level $\ell$.
We quantize reference vertices into unique occupied cells at the finest resolution and construct the octree by collecting their ancestors.
Each parent $i$ has eight children indexed by $o=4x+2y+z$, where $x,y,z\in\{0,1\}$.
Its target occupancy belongs to the nonempty support
\begin{equation}
    O_i\in\mathcal O=\{0,1\}^{8}\setminus\{\mathbf 0\},
    \qquad |\mathcal O|=255.
    \label{eq:occupancy_support}
\end{equation}
Let $h_i\in\mathbb R^{d_b}$ denote the parent feature summarizing the current frontier state and geometric condition; architecture dimensions are defined in \cref{sec:network_architecture}.
We suppress $i$ and $\ell$ when discussing a fixed parent and level.

\paragraph{Spatial decisions and complete history.}
We recursively bisect a parent along the X, Y, and Z axes, forming seven internal regions and eight child-cell leaves.
For every active region, a decision $s_j\in\{10,01,11\}$ indicates whether only its left half, only its right half, or both halves are nonempty.
An inactive region is deterministically empty.
The seven regions are visited in breadth-first order: the parent, its two halves, and its four quarter regions.
Every valid occupancy pattern corresponds to a unique sequence of active decisions, with three to seven decisions per pattern.
This spatial decision hierarchy operates within a parent, while octree refinement advances between resolution levels.

A tree state $(j,s_{<j})$ pairs the current slot with all preceding decisions, including siblings and other subtrees.
Known empty regions use $\varnothing$; current and unvisited slots use $?$.
For example, occupancy at children 0 and 7 yields
$[11,10,01,10,\varnothing,\varnothing,01]$;
the input for decision $j=3$ is $[11,10,01,?,?,?,?]$.
Each decision can thus depend on earlier spatial choices beyond its ancestor path.

\paragraph{Joint occupancy posterior.}
Conditional probability products are a standard basis for tractable distribution estimation~\cite{made2015}; here the variables are nonempty spatial-region decisions over the valid occupancy support.
Let $u=s_{<j}$ denote the current history, $g_\theta(h,j,u)\in\mathbb R^{d_o}$ the learned logits for branches $a\in\{10,01,11\}$, and $n(ua)$ the number of valid complete patterns consistent with taking branch $a$.
The shared conditional MLP applies \emph{branch normalization} in its output stage by adding the fixed log completion count to the learned branch logits and then applying softmax:
\begin{equation}
    \pi_\theta(a\mid h,j,u)
    =\operatorname{softmax}_{a}\!\left[g_\theta(h,j,u)+\log n(ua)\right].
    \label{eq:branch_probability}
\end{equation}
The fixed completion-count offset gives $p_0(O)=1/255$ when the learned logits are zero, avoiding the bias toward sparse patterns induced by uniform three-way branching.
Let $A(O)$ contain the active slots of pattern $O$.
\emph{Posterior composition} defines its joint probability as
\begin{equation}
    p_\theta(O\mid h)
    =\prod_{j\in A(O)}
    \pi_\theta\!\left(s_j(O)\mid h,j,s_{<j}(O)\right).
    \label{eq:pattern_posterior}
\end{equation}
The pattern--sequence bijection and normalized branch probabilities imply
$\sum_{O\in\mathcal O}p_\theta(O\mid h)=1$.
Proofs of normalization and uniform initialization are given in Suppl.~\cref{sec:sequence_proof,sec:initialization_proof}.

\subsection{Network Architecture}
\label{sec:network_architecture}

The occupancy posterior network pairs a shared DiT backbone with a shared spatial conditional MLP that includes branch normalization.

\paragraph{Point-cloud condition.}
A point-cloud encoder aggregates Fourier-encoded coordinates and surface normals through cross-attention~\cite{attention2017}.
Its spatial queries are centers of coarse cells occupied by the input points, with at most 2,048 queries.
An eight-block, 1,024-channel Transformer with 16 attention heads refines these features into normalized context tokens for the vertex generator.
The encoder is initialized from pretrained weights and jointly optimized with the generator.

\paragraph{Global frontier backbone.}
A shared DiT~\cite{dit2022} supplies global context to each parent through self-attention over the entire frontier and cross-attention to the point-cloud tokens.
It uses $N_{\mathrm{DiT}}$ $(=24)$ blocks of width $d_b$ $(=768)$ with eight attention heads.
Each parent token combines its projected eight-channel state $X_{t,i}$ with depth and address embeddings; the address encodes the root-to-parent octant path.
Spatial attention uses 3D rotary position encoding, while time and an optional vertex-count condition $\nu$ enter through adaptive modulation.
Embedding $\log\nu$, or a fixed null value when absent, provides soft control over the final vertex count.
Writing $X_t$ for all parent states, the normalized and modulated output feature is
\begin{equation}
    S=(X_t,t,C,\mathcal F_\ell,\nu),
    \qquad h_i=f_\theta(S)_i\in\mathbb R^{d_b}.
    \label{eq:backbone_context}
\end{equation}

\paragraph{Shared spatial conditional MLP.}
\emph{Tree-state embedding} encodes the partially resolved local spatial tree and the current decision slot $j$.
The five history symbols $\{\varnothing,10,01,11,?\}$ share an embedding table of width $d_e$ $(=16)$; the seven decision slots use a separate table of the same width.
Concatenating the seven history embeddings and the current slot embedding gives the tree-state embedding $e=e(j,s_{<j})\in\mathbb R^{d_c}$, where $d_c=(7+1)d_e$ $(=128)$.
The MLP's learned layers have $N_{\mathrm{MLP}}$ $(=2)$ hidden layers of width $d_h$ $(=128)$ and a final logits projection of width $d_o$ $(=3)$ corresponding to the nonempty branch choices.
We project $h_i$ and $e$ into the hidden space and compute
\begin{equation}
    \begin{aligned}
    r_1&=\operatorname{SiLU}(W_hh_i+W_ee+b_1),\\
    r_k&=\operatorname{SiLU}(W_kr_{k-1}+b_k),\\
    g_\theta(h_i,j,s_{<j})&=W_or_{N_{\mathrm{MLP}}}+b_o.
    \end{aligned}
    \label{eq:shared_head}
\end{equation}
All hidden states $r_k$ have width $d_h$; the recurrence applies for $2\le k\le N_{\mathrm{MLP}}$.
In \cref{fig:method}, $k$ indexes the repeated SiLU--Linear sequence: its output width is $d_k=d_h$ for $1\le k<N_{\mathrm{MLP}}$ and $d_{N_{\mathrm{MLP}}}=d_o$ at the final iteration.
The MLP's fixed output stage applies branch normalization in \cref{eq:branch_probability}, converting its internal logits $g_\theta$ into conditional branch probabilities $\pi_\theta$.
The final learned linear layer is initialized to zero, yielding uniform pattern probabilities through this output stage.
This MLP and the backbone are shared across parents and octree levels; the MLP is also shared across all seven decision slots.
The tree-state embedding enters only the MLP, allowing different candidate tree states to reuse the same backbone feature.

\subsection{Posterior-Driven Flow Matching}
\label{sec:posterior_flow}
\label{sec:occupancy_path}

\paragraph{Continuous path and likelihood training.}
We embed occupancy as $Z_i=2O_i-\mathbf 1$ and interpolate from independent Gaussian noise,
\begin{equation}
    X_{t,i}=(1-t)X_{0,i}+tZ_i,
    \qquad X_{0,i}\sim\mathcal N(\mathbf 0,I_8).
    \label{eq:occupancy_path}
\end{equation}
This follows the continuous flow-matching setting~\cite{flowmatching2022,rectifiedflow2022} with a finite occupancy endpoint.
At each sampled time, the network learns the occupancy posterior by minimizing
\begin{equation}
    \begin{aligned}
    \mathcal L_{\mathrm{post}}
    &=\mathbb E[-\log p_\theta(O\mid h)]\\
    &=\mathbb E\!\left[-\sum_{j\in A(O)}
       \log\pi_\theta(s_j\mid h,j,s_{<j})\right].
    \end{aligned}
    \label{eq:posterior_nll}
\end{equation}
The expectation samples objects, refinement levels, noise, and time; parent losses are averaged within each object.
Active decision losses are summed without dividing by path length, preserving the complete-pattern likelihood.
Training evaluates the target decisions with their observed histories, whereas inference evaluates candidate tree states to obtain the full distribution.

\paragraph{Posterior mean and continuous velocity.}
The normalized posterior gives an exactly computable endpoint mean,
\begin{equation}
    m_\theta(h)=\sum_{O\in\mathcal O}p_\theta(O\mid h)(2O-\mathbf 1).
    \label{eq:posterior_mean}
\end{equation}
For the path in \cref{eq:occupancy_path}, conditioning on an endpoint gives
$\dot X_{t,i}=(Z_i-X_{t,i})/(1-t)$.
Consequently, its marginal flow velocity is the posterior expectation~\cite{flowmatching2022},
\begin{equation}
    \begin{aligned}
    v_i^*(S)
    &=\mathbb E\!\left[\frac{Z_i-X_{t,i}}{1-t}
        \;\middle|\;S\right]\\
    &=\frac{\mathbb E[Z_i\mid S]-X_{t,i}}{1-t}.
    \end{aligned}
    \label{eq:posterior_velocity_identity}
\end{equation}
TOPFlow substitutes the learned mean and integrates
\begin{equation}
    \frac{\mathrm dX_{t,i}}{\mathrm dt}
    =v_{\theta,i}=\frac{m_\theta(h_i)-X_{t,i}}{1-t},
    \qquad t<1.
    \label{eq:learned_velocity}
\end{equation}
The likelihood objective learns the endpoint distribution, and its mean supplies the continuous update without an additional velocity-regression loss.
When the true conditional posterior is representable, the population likelihood optimum recovers its mean and hence the marginal velocity (Suppl.~\cref{sec:proper_likelihood,sec:velocity_proof}).
Endpoint MSE shares this ideal Bayes mean; finite-budget improvements are evaluated empirically.

\paragraph{Coarse-to-fine sampling.}
With shared backbone context, the MLP batches 127 candidate tree states to obtain branch probabilities for all 255 patterns.
Each conditional backbone is evaluated once per integration step.
Starting from the occupied root, we update all parents' continuous states in parallel within a level.
At the end of a level, positive state entries identify occupied children; if all entries are nonpositive, we retain the largest one.
These cells form the next frontier.
The final occupied-cell centers become vertices, which are connected into a mesh using Nexus's Spacetime Interval architecture.
The finite sum in \cref{eq:posterior_mean} is exact for the learned local posterior; posterior estimation and numerical integration remain approximate.

\subsection{Count and Multiscale Layout Control}
\label{sec:temperature_control}

For a fixed posterior $P(O)=p_\theta(O\mid h)$, let $K=\sum_oO_o$ be the occupied-child count and $L=\sum_{o=0}^{3}O_o$ the count in the X-axis left half.
The distribution factors exactly as
\begin{equation}
    P(O)=P(K)\,P(L\mid K)\,P(O\mid K,L).
    \label{eq:occupancy_factorization}
\end{equation}
These factors respectively describe total count, coarse allocation, and the fine layout at a fixed allocation.
All are obtained by summing and normalizing the same pattern posterior.
For a categorical distribution $r$, define its temperature transform by
\begin{equation}
    \mathcal T_T[r](a)=\frac{r(a)^{1/T}}{\sum_b r(b)^{1/T}},
    \qquad T>0.
    \label{eq:temperature_transform}
\end{equation}
We independently control the three factors:
\begin{equation}
    \begin{aligned}
    Q_{\boldsymbol T}(O)
    ={}&\mathcal T_{T_K}[P(K)]\,
         \mathcal T_{T_C}[P(L\mid K)]\\
       &\cdot\mathcal T_{T_F}[P(O\mid K,L)].
    \end{aligned}
    \label{eq:factor_temperatures}
\end{equation}
Each factor is normalized on its conditional support, so $Q_{\boldsymbol T}$ is normalized and $Q_{(1,1,1)}=P$.
The controlled mean replaces $P$ with $Q_{\boldsymbol T}$ in \cref{eq:posterior_mean}.
Temperature scaling has a classical role in probability calibration~\cite{guo2017calibration}; here it provides inference control over geometric occupancy.

At $T=0$, we use the limit of \cref{eq:temperature_transform}, which retains uniform mass over tied maximizers.
With unique conditional maxima, $(0,0,0)$ selects a count, then an allocation, then a layout; this conditional selection generally differs from the full-pattern MAP.
At a fixed posterior, changing $T_F$ preserves the count and coarse-allocation marginals, and changing $T_C$ preserves the count marginal.
These temperatures control local occupancy factors, while $\nu$ is a condition on the final vertex count; changes to the flow trajectory can affect subsequent geometry.
Their quality effects and the default temperatures are established through experiments.

For classifier-free guidance~\cite{cfg2022}, we apply the same temperature transform to posteriors with and without point-cloud conditioning, keeping the vertex-count condition fixed.
Their means are $m_c$ and $m_u$, respectively.
The guided mean $m_u+w(m_c-m_u)$, with guidance scale $w$, replaces $m_\theta$ in \cref{eq:learned_velocity} to compute the flow velocity.

\section{Experiments}
\label{sec:experiments}

\subsection{Implementation Details}

\paragraph{Training and inference.}
\textbf{Training.}
We train TOPFlow on approximately 1.17M filtered Objaverse/Objaverse-XL artistic meshes~\cite{objaverse2023,objaversexl2023}, normalized within $[-1,1]^3$ and quantized onto a $1024^3$ octree ($L=10$).
Training takes approximately six days for 350k updates on 8 NVIDIA H20 GPUs, using AdamW~\cite{adamw2017}, a learning rate of $10^{-4}$, a global batch size of 128, and 2,000 warm-up updates followed by cosine decay.
We sample one refinement level per object and jointly optimize the point-cloud conditioner.
\textbf{Inference.}
We use exponential-moving-average weights, 20 Euler updates per level, and point-cloud classifier-free guidance~\cite{cfg2022} with scale 2.
Inputs contain 16,384 or 40,960 surface points with normals (16K/40K), without retraining.
The vertex-count condition is absent on Toys4K and Objaverse and set to 4,000 on DenseMesh; default temperatures are $(T_K,T_C,T_F)=(0,0,0)$.

\paragraph{Benchmarks and Metrics.}
We evaluate on Toys4K~\cite{toys4k2021}, Objaverse~\cite{objaverse2023}, and \densemeshset{}, with 1,068, 500, and 222 test objects, respectively, excluded from training by UID.
Toys4K and Objaverse contain existing mesh assets; \densemeshset{} contains dense meshes synthesized by large-scale 3D generative models~\cite{ultrashape2025,lattice2025,trellis22025,hunyuan3d2025hunyuan3d21,xiang2025trellis}, testing detail preservation in compact mesh generation.
Their surfaces provide input conditions and evaluation references.
\textbf{Metrics.}
Complete meshes are evaluated using Chamfer distance (CD, $\times10^3$), Hausdorff distance (HD, $\times10^2$), and normal consistency (NC).
For vertex generation, vertex-to-surface distance (V2S, $\times10^3$) measures surface fidelity, density-aware Chamfer distance (DCD)~\cite{dcd2021} measures vertex-layout agreement while penalizing local concentration, and surface F-score (F, \%) indirectly assesses compatibility with the connectivity model's learned vertex distribution through generated mesh quality.
F-score uses $\tau=1/128$, equivalent to four finest-level voxel widths.
Metric formulas are given in Suppl.~\cref{sec:metric_definitions}.

\begin{table}[!t]
\centering
\small
\caption{\textbf{Final mesh quality on Toys4K and Objaverse.} $\dagger$Controlled reimplementation; $\ddagger$image + sparse-voxel input.}
\label{tab:mesh}
\setlength{\tabcolsep}{3pt}
\renewcommand{\arraystretch}{1.06}
\resizebox{\columnwidth}{!}{%
\begin{tabular}{@{}lrrrrrr@{}}
\toprule
 & \multicolumn{3}{c}{\textbf{Toys4K}} & \multicolumn{3}{c}{\textbf{Objaverse}} \\
\cmidrule(lr){2-4}\cmidrule(lr){5-7}
Method & CD$\downarrow$ & HD$\downarrow$ & NC$\uparrow$ & CD$\downarrow$ & HD$\downarrow$ & NC$\uparrow$ \\
\midrule
\multicolumn{7}{l}{\textit{Autoregressive}} \\
EdgeRunner~\cite{edgerunner2024} & 15.26 & 17.12 & 0.905 & 18.92 & 20.92 & 0.894 \\
TreeMeshGPT~\cite{treemeshgpt2025} & 10.51 & 13.91 & 0.914 & 12.04 & 17.51 & 0.904 \\
BPT~\cite{bpt2024} & 6.36 & 8.26 & 0.931 & 6.67 & 11.04 & 0.925 \\
DeepMesh~\cite{deepmesh2025} & 8.02 & 11.47 & 0.939 & 8.28 & 14.55 & 0.937 \\
FastMesh~\cite{fastmesh2025} & 7.86 & 9.08 & 0.899 & 12.83 & 14.54 & 0.883 \\
MeshRipple~\cite{meshripple2025} & 7.26 & 10.20 & 0.919 & 8.74 & 14.62 & 0.912 \\
\midrule
\multicolumn{7}{l}{\textit{Diffusion}} \\
MeshFlow~\cite{meshflowmeta2026} & 9.75 & 6.45 & 0.893 & 13.45 & 8.35 & 0.878 \\
LATO.2~\cite{lato22026}$^{\ddagger}$ & 5.10 & 6.73 & 0.897 & 6.37 & 8.12 & 0.885 \\
Nexus (16K)$^{\dagger}$ & 2.67 & 8.47 & 0.961 & 3.97 & 11.42 & 0.955 \\
Nexus (40K)$^{\dagger}$ & 2.24 & 7.32 & 0.967 & 2.99 & 9.25 & 0.960 \\
\rowcolor{papergray} TOPFlow (16K) & \textbf{1.66} & \textbf{4.35} & \underline{0.976} & \underline{2.40} & \underline{6.89} & \underline{0.972} \\
\rowcolor{papergray} TOPFlow (40K) & \underline{2.07} & \underline{4.71} & \textbf{0.977} & \textbf{2.23} & \textbf{6.84} & \textbf{0.972} \\
\bottomrule
\end{tabular}
}
\end{table}

\begin{table}[!t]
\centering
\small
\caption{\textbf{Vertex generation quality on Toys4K and Objaverse.} $\dagger$Controlled reimplementation.}
\label{tab:vertex}
\setlength{\tabcolsep}{3pt}
\renewcommand{\arraystretch}{1.06}
\resizebox{\columnwidth}{!}{%
\begin{tabular}{@{}lrrrrrr@{}}
\toprule
 & \multicolumn{3}{c}{\textbf{Toys4K}} & \multicolumn{3}{c}{\textbf{Objaverse}} \\
\cmidrule(lr){2-4}\cmidrule(lr){5-7}
Method & V2S$\downarrow$ & DCD$\downarrow$ & F$\uparrow$ & V2S$\downarrow$ & DCD$\downarrow$ & F$\uparrow$ \\
\midrule
Nexus (16K)$^{\dagger}$ & 1.52 & 0.339 & 94.92 & 1.47 & 0.370 & 93.64 \\
Nexus (40K)$^{\dagger}$ & 1.44 & 0.317 & 96.19 & 1.42 & 0.341 & 95.14 \\
\rowcolor{papergray} TOPFlow (16K) & \underline{1.43} & \underline{0.277} & \underline{97.87} & \underline{1.34} & \underline{0.315} & \underline{97.06} \\
\rowcolor{papergray} TOPFlow (40K) & \textbf{1.34} & \textbf{0.262} & \textbf{97.88} & \textbf{1.25} & \textbf{0.292} & \textbf{97.20} \\
\bottomrule
\end{tabular}
}
\end{table}

\begin{figure*}[!t]
    \centering
    \includegraphics[width=\textwidth]{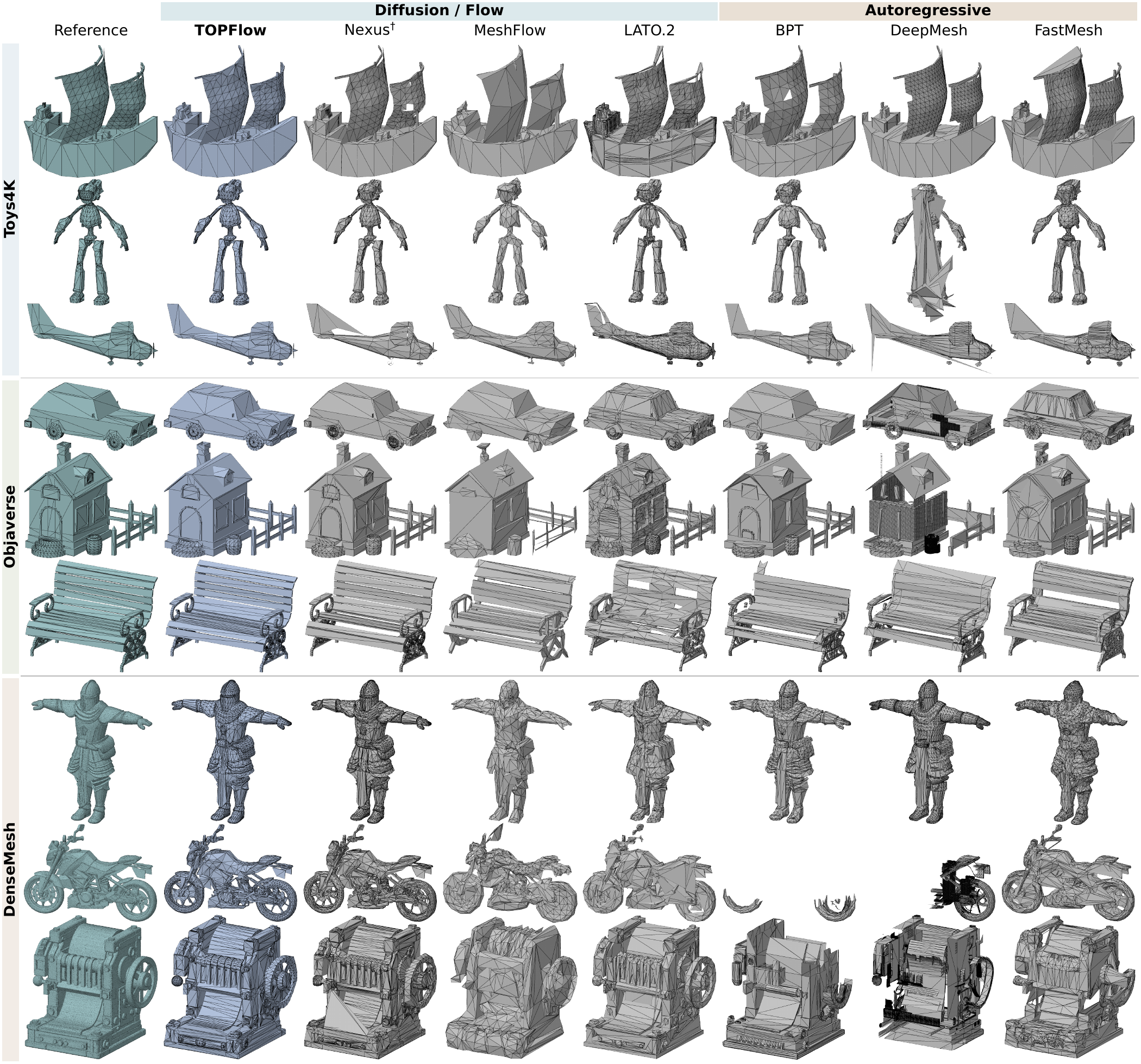}
    \caption{\textbf{Qualitative mesh comparison across Toys4K, Objaverse, and DenseMesh.}
    $^{\dagger}$Controlled reimplementation.}
    \label{fig:qualitative}
\end{figure*}

\begin{figure*}[!t]
    \centering
    \includegraphics[width=\textwidth]{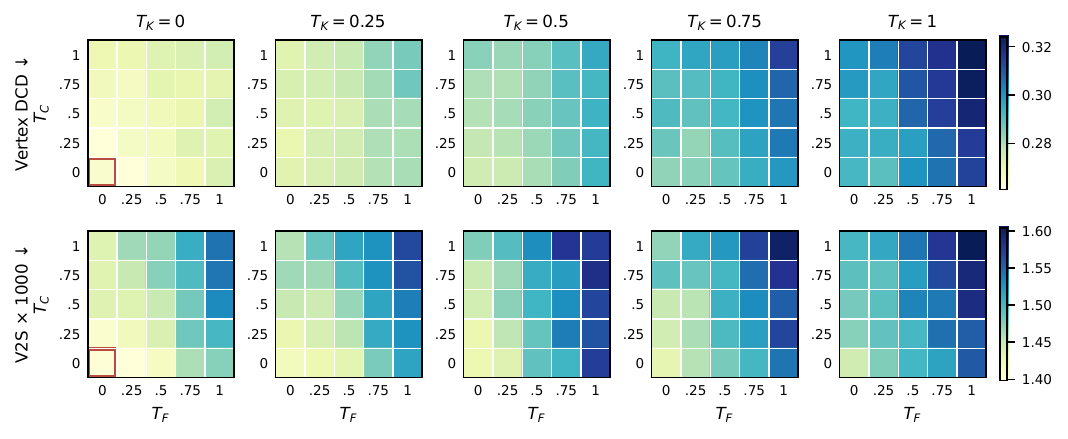}
    \par\vspace{1mm}
    \includegraphics[width=\textwidth]{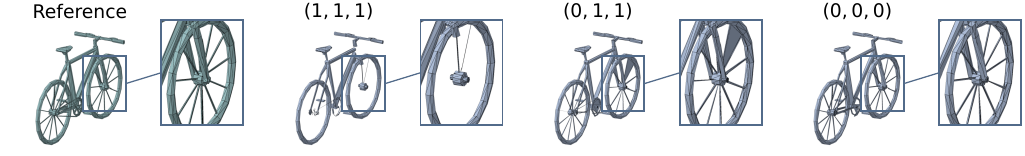}
    \caption{\textbf{Temperature control on Toys4K.}
    Each panel shows a heatmap over $T_C$ and $T_F$ at a fixed $T_K$.
    Panels for the same metric share a color scale; lower values are better.
    Outlined cells mark $000$.
    Matched wheel close-ups compare $111$, $011$, and $000$, illustrating improved spoke structure.}
    \label{fig:temperature}
\end{figure*}

\subsection{Mesh and Vertex Quality}

\paragraph{Quantitative results.}
\cref{tab:mesh,tab:vertex} show TOPFlow achieving the best mesh CD, HD, and NC among the compared methods, and improving all vertex-quality metrics over Nexus on Toys4K and Objaverse.
MeshFlow~\cite{meshflowmeta2026} has relatively low HD but higher CD and lower NC, a trade-off that may partly reflect reconstruction error from its compressed MeshVAE representation.
Nexus~\cite{nexus2026} directly predicts continuous velocities for octree occupancy, whereas TOPFlow learns conditional occupancy decisions before computing the velocity; the lower DCD is consistent with better modeling of local vertex arrangements.
With 60\% fewer input points (16K vs.\ 40K), TOPFlow outperforms Nexus on all three vertex metrics, reducing DCD by 12.6\% on Toys4K and 7.6\% on Objaverse.
DenseMesh quantitative results are reported in Suppl.~\cref{sec:dense_results}.

\paragraph{Qualitative comparisons.}
\cref{fig:qualitative} compares surface completeness and local detail across the three datasets.
Autoregressive methods~\cite{bpt2024,deepmesh2025,fastmesh2025} remain susceptible to completeness failures during sequential mesh generation, as illustrated by missing or fragmented motorcycle parts.
MeshFlow captures the overall silhouettes but simplifies the car and machinery details.
Compared with Nexus, which leaves holes in the sail, TOPFlow preserves the sail surface and fine structures such as fence elements and motorcycle wheels more faithfully.

\subsection{Ablation Studies}

\begin{table}[t]
\centering\footnotesize
\caption{\textbf{Posterior parameterization and temperature ablation on Toys4K.} Parameter counts cover the vertex denoiser. $\dagger$Controlled reimplementation.}
\label{tab:ablation}
\setlength{\tabcolsep}{2pt}
\resizebox{\columnwidth}{!}{%
\begin{tabular}{@{}llrrrrr@{}}
\toprule
Model & $\Delta$ & \shortstack{Params\\(M)} & V2S$\downarrow$ & DCD$\downarrow$ & F$\uparrow$ & \shortstack{Mesh\\CD$\downarrow$} \\
\midrule
Nexus$^{\dagger}$ & -- & 366.913 & \underline{1.44} & 0.317 & 96.19 & 2.24 \\
Flat Linear ($T=1$) & Discrete & 367.103 & 1.60 & 0.318 & 97.24 & 2.49 \\
Flat MLP ($T=1$) & Discrete & 367.055 & 1.64 & 0.317 & 97.29 & 2.48 \\
TOPFlow ($111$) & \shortstack[l]{Discrete\\+ spatial} & 367.039 & 1.55 & \underline{0.314} & \underline{97.63} & \underline{2.14} \\
\rowcolor{papergray} TOPFlow ($000$) & \shortstack[l]{Discrete\\+ spatial\\+ temp.} & 367.039 & \textbf{1.34} & \textbf{0.262} & \textbf{97.88} & \textbf{2.07} \\
\bottomrule\end{tabular}
}
\end{table}

\paragraph{Posterior parameterization.}
\cref{tab:ablation} examines the progression from continuous endpoint supervision to discrete posterior supervision, spatial conditional factorization, and temperature control on Toys4K with 40K inputs.
Both direct heads replace endpoint regression with a 255-pattern posterior whose mean drives continuous flow sampling.
Flat Linear maps the backbone feature directly to 255 logits with a single linear layer; Flat MLP uses two 128-channel hidden layers with SiLU activations before the 255-logit output layer.
This discrete supervision improves surface F-score over Nexus, but yields little change in DCD and higher V2S and mesh CD, indicating limited gains from supervision alone.
Adding the spatial conditional posterior in TOPFlow at unit temperatures reduces DCD by 1.0\% and mesh CD by 13.8\% relative to Flat MLP.
Applying zero temperatures further reduces DCD by 16.6\% and mesh CD by 3.1\%, with improvements in V2S and surface F-score.
The strongest vertex-layout gains thus emerge when spatial posterior modeling is combined with temperature control.
TOPFlow's spatial conditional posterior adds only \addedvertexpercent{} vertex-denoiser parameters relative to Nexus.

\paragraph{Temperature control.}
To assess the individual and joint effects of temperature control and identify suitable settings, we ablate TOPFlow's three factors: $T_K$ controls the occupied-child count, $T_C$ its coarse spatial allocation, and $T_F$ the fine layout within that allocation.
Using 40K inputs and fixed model weights, we test all 125 combinations of $\{0,0.25,0.5,0.75,1\}$ on 128 Toys4K objects; \cref{fig:temperature} shows the resulting DCD and V2S heatmaps.
The heatmaps show an overall trend toward lower DCD and V2S at lower temperatures, with favorable settings concentrated near $000$.
Reducing $T_K$ has a particularly clear effect, while low coarse and fine temperatures further improve vertex layouts.
Zero temperature concentrates each conditional factor on its most probable choices, reducing averaging across competing occupancy patterns in the posterior mean.
The matched wheel close-ups in \cref{fig:temperature} illustrate this effect: reducing count temperature from $111$ to $011$ restores missing spokes, and $000$ further improves their arrangement.
Detailed numerical results, independent validation, and further analysis are provided in Suppl.~\cref{sec:temperature_report}.

\section{Conclusion}

TOPFlow learns a tree-structured occupancy posterior for mesh vertex generation.
A shared spatial conditional head learns joint child occupancy from complete histories via exact likelihood; its mean drives continuous flow matching.
Separate count, coarse, and fine temperatures control inference.
TOPFlow achieves state-of-the-art mesh generation quality on Toys4K, Objaverse, and DenseMesh.

\paragraph{Limitations and Future Work.}
TOPFlow refines one octree level at a time, preventing spatial regions from advancing across depths at different rates.
Adaptive refinement across levels could provide greater flexibility in allocating detail.
Continuous flow sampling also requires repeated network evaluations for numerical integration.
Future work could explore fewer-step sampling to reduce this computational cost while preserving mesh quality.

\FloatBarrier
{
    \small
    \bibliographystyle{ieeenat_fullname}
    \bibliography{main}
}

\renewcommand{\maketitlesupplementary}{}
\clearpage
\appendix
\maketitlesupplementary

\section{Proofs for the Occupancy Posterior}
\label{sec:posterior_proofs}

Using the notation of \cref{sec:method} of the main paper, we first establish normalization and uniform initialization of the occupancy posterior, then connect likelihood training to posterior-mean accuracy and the marginal flow velocity.

\subsection{Valid Sequences and Normalization}
\label{sec:sequence_proof}

A nonempty occupancy pattern uniquely determines whether each visited region's left half, right half, or both halves are nonempty.
Conversely, a valid decision sequence activates at least one child of every active region; after the X, Y, and Z splits, its active leaves determine a unique nonempty pattern.
These constructions are inverse, so summing over valid decision sequences is equivalent to summing over the 255 nonempty occupancy patterns.
Conditioned on the backbone context $h$ and the tree-state embedding $e(j,u)$ from main \cref{sec:network_architecture}, the shared conditional MLP's branch normalization in main \cref{eq:branch_probability} ensures $\sum_a\pi_\theta(a\mid h,j,u)=1$ at every active slot.
To make the normalization explicit, let $W(u)$ be the summed probability of all continuations after a valid prefix $u$.
A completed prefix has $W(u)=1$, since no decisions remain.
For an active next slot, assume that each continuation $ua$ is normalized, so $W(ua)=1$.
Combining this induction hypothesis with normalized branch probabilities gives
\begin{equation}
 W(u)=\sum_{a\in\{10,01,11\}}\pi_\theta(a\mid h,j,u)W(ua)=1.
 \label{eq:prefix_normalization}
\end{equation}
An inactive slot has one empty continuation with probability one.
The induction conditions on each specific prefix, so dependencies on siblings and earlier subtrees do not affect normalization.
Applying the induction at the initial prefix establishes normalization of posterior composition over all 255 patterns.
For a target pattern, taking the negative logarithm of the composed probability converts its branch-probability product into the sum of active decision losses in main \cref{eq:posterior_nll}.
Dividing this sum by the number of active decisions would instead weight each pattern's likelihood contribution inversely by its path length.

\subsection{Uniform Initialization}
\label{sec:initialization_proof}

The MLP's learned $d_h\to d_o$ projection in main \cref{eq:shared_head} is zero-initialized, giving zero internal logits for every backbone context and tree-state embedding.
Let $n(u)$ be the number of complete patterns compatible with prefix $u$.
Its active continuations partition that set, so $n(u)=\sum_a n(ua)$.
Its fixed branch normalization in main \cref{eq:branch_probability} then gives $\pi_0(a\mid u)=n(ua)/n(u)$.
Along a pattern's path, each numerator is the completion count used in the next denominator; inactive slots leave this count unchanged.
Posterior composition therefore cancels all intermediate completion counts:
\begin{equation}
 p_0(O)=\prod_{j\in A(O)}\frac{n(u_{j+1})}{n(u_j)}
       =\frac{1}{255}.
\end{equation}
The remaining denominator is 255, the number of completions at the initial prefix; the remaining numerator is one, since a completed prefix specifies a single pattern.
For example, the root continuations $10$, $01$, and $11$ admit 15, 15, and 225 completions.
Assigning each root branch probability $1/3$ would give these differently sized pattern groups equal total mass, producing a nonuniform pattern distribution.
The fixed offsets account for these completion counts at initialization.
Learned logits can cancel or modify the offsets, so this initialization does not restrict the positive conditional distributions the MLP can represent.

\subsection{Likelihood and Mean Error}
\label{sec:proper_likelihood}
\label{sec:mean_error_proof}

Using $S=(X_t,t,C,\mathcal F_\ell,\nu)$ and $h(S)\in\mathbb R^{d_b}$ from main \cref{eq:backbone_context}, let $p^*(O\mid S)$ be the true local posterior and $q(O\mid S)=p_\theta(O\mid h(S))$ the learned posterior.
The logarithmic scoring identity~\cite{proper2007} is
\begin{equation}
 \mathbb E_{p^*}[-\log q]=H(p^*)+\mathrm{KL}(p^*\|q).
 \label{eq:conditional_logloss}
\end{equation}
Adding and subtracting $\mathbb E_{p^*}[\log p^*]$ proves the identity.
Posterior composition and the logarithm of a product make the summed $d_o$-class branch cross-entropies equal to this complete-pattern objective.
Since $H(p^*)$ is independent of the model and KL divergence is nonnegative, the population objective is minimized at $q=p^*$ whenever the true posterior is representable.
The same decomposition holds after averaging over the object and parent weights used in training.

For distributions $p,q$ on the valid patterns, define $\mu_p=\mathbb E_p[2O-\mathbf1]$ and $\mathrm{TV}(p,q)=\frac12\sum_O|p(O)-q(O)|$.
Each endpoint coordinate is bounded by one, hence $|\mu_{p,o}-\mu_{q,o}|\le2\mathrm{TV}(p,q)$.
Indeed, the coordinate difference is $\sum_O(p(O)-q(O))(2O_o-1)$, whose absolute value is bounded by $\sum_O|p(O)-q(O)|$.
Summing eight squared bounds and applying Pinsker's inequality~\cite{coverthomas2006} yields
\begin{equation}
 \|\mu_p-\mu_q\|_2^2
 \le32\mathrm{TV}(p,q)^2
 \le16\mathrm{KL}(p\|q).
 \label{eq:mean_kl_bound}
\end{equation}
Taking $p=p^*$ makes KL divergence the excess population log loss, so this bound connects likelihood accuracy to endpoint-mean accuracy.
The converse does not follow: different joint distributions can share the same mean.
An ideal endpoint-MSE predictor recovers that same Bayes mean, so the bound alone does not establish an advantage over MSE; improvements under finite training budgets require empirical evaluation.

\subsection{Marginal Flow Velocity}
\label{sec:velocity_proof}

Fix the frontier, point-cloud condition, and optional count condition, and stack all parents' endpoints into $Z$.
From main \cref{eq:occupancy_path}, $\dot X_t=(Z-X_t)/(1-t)$ for $t<1$.
Here $X_0$ and $Z$ are fixed along each sampled path; the derivative is $Z-X_0$, and the displayed form follows by solving the interpolation for $X_0$.
For a smooth test function $\varphi$, differentiation and conditional expectation give
\begin{equation}
 \begin{aligned}
 \frac{\mathrm d}{\mathrm dt}\mathbb E[\varphi(X_t)]
 &=\mathbb E[\nabla\varphi(X_t)^\top\dot X_t]\\
 &=\mathbb E[\nabla\varphi(X_t)^\top
                    \mathbb E[\dot X_t\mid S]].
 \end{aligned}
\end{equation}
The last equality replaces each sampled path velocity with its conditional expectation at the current state.
It therefore establishes the marginal velocity in main \cref{eq:posterior_velocity_identity} through the weak continuity equation~\cite{flowmatching2022}.
The local posterior conditions on the whole frontier state, so no independence between parent endpoints is required.
The learned velocity equals this field when its posterior mean is correct.
At $t\le1-\delta$, \cref{eq:mean_kl_bound} bounds local squared velocity error by $16\mathrm{KL}(p^*\|q)/\delta^2$.
This error bound grows as the endpoint is approached and does not by itself guarantee the accuracy of a finitely integrated trajectory.
Temperature control and guidance further modify the posterior mean, and therefore the velocity field to which the ideal identity applies.

\section{Properties of Temperature Control}
\label{sec:temperature_proofs}

\subsection{Normalization and Preserved Marginals}

Use the factors defined in main \cref{eq:occupancy_factorization,eq:factor_temperatures}, with conditionals restricted to positive-mass groups.
Let $\mathcal O_{k,l}$ contain patterns with total count $k$ and left-half count $l$.
The grouped probabilities and conditional layout are
\begin{equation}
 \begin{aligned}
 P(k,l)&=\sum_{O\in\mathcal O_{k,l}}P(O),\quad P(k)=\sum_l P(k,l),\\
 P(l\mid k)&=\frac{P(k,l)}{P(k)},\quad
 P(O\mid k,l)=\frac{P(O)}{P(k,l)}.
 \end{aligned}
 \label{eq:temperature_groups}
\end{equation}
Zero-mass groups contribute nothing and need no conditional definition.
In the following sums, $Q(K)$, $Q(L\mid K)$, and $Q(O\mid K,L)$ denote the three temperature-transformed factors in main \cref{eq:factor_temperatures}.
For each occupied count $K$, the feasible allocations satisfy $\max(0,K-4)\le L\le\min(4,K)$.
Successively summing over patterns at fixed $(K,L)$, allocations at fixed $K$, and counts gives
\begin{equation}
 \sum_O Q_{\boldsymbol T}(O)=\sum_K Q(K)\sum_L Q(L\mid K)
                         \sum_{O:K,L}Q(O\mid K,L)=1.
\end{equation}
At unit temperatures, each transformed factor is unchanged and the chain rule recovers $P$.
The controlled marginals satisfy
\begin{equation}
 \begin{aligned}
 Q(K)&=\mathcal T_{T_K}[P(K)],\\
 Q(K,L)&=Q(K)\mathcal T_{T_C}[P(L\mid K)].
 \end{aligned}
\end{equation}
Hence changing $T_F$ only redistributes probability among layouts with the same $(K,L)$, preserving $Q(K,L)$.
Changing $T_C$ redistributes probability among allocations within each count, preserving $Q(K)$ at fixed $T_K$.
These properties concern a fixed posterior at one integration step; the resulting change in the flow state can affect later posteriors and generated geometry.

\subsection{Zero-Temperature Limit}

For a categorical distribution $r$, let $M=\max_a r(a)>0$ and $A=\{a:r(a)=M\}$.
Dividing numerator and denominator of main \cref{eq:temperature_transform} by $M^{1/T}$ shows that nonmaximal terms vanish as $T\to0^+$:
\begin{equation}
 \lim_{T\to0^+}\mathcal T_T[r](a)=\frac{\mathbf1[a\in A]}{|A|}.
\end{equation}
Retaining all tied maxima therefore preserves normalization.
With unique conditional maxima, zero temperatures select count, then allocation conditioned on that count, then pattern conditioned on both.
The most probable count can contain several patterns, so this conditional selection need not recover the single pattern with the largest joint probability (full-pattern MAP).
For example, assign probabilities $0.3,0.3,0.4$ to $e_0,e_1,e_0+e_1$, where $e_o$ occupies only child $o$.
The count marginal assigns mass $0.6$ to $K=1$ and $0.4$ to $K=2$, so zero count temperature selects $K=1$.
Children 0 and 1 lie in the same X-axis half, and subsequent concentration retains their two tied single-child patterns.
Full-pattern MAP instead selects $e_0+e_1$, whose individual probability is $0.4$.
Thus zero temperature expresses a preference at each conditional factor; whether this produces better geometry is assessed empirically.

\section{Additional Quantitative Results}
\label{sec:evaluation_details}

\subsection{Metric Definitions}
\label{sec:metric_definitions}

Let $V,V^*$ be unique generated and reference vertices, and $S,S^*$ their mesh surfaces, normalized to a longest reference bounding-box extent of 1.98.
Define $d(x,S)=\inf_{y\in S}\|x-y\|_2$.
Only exact-position duplicate vertices are removed.
Mesh distances query continuous triangles: samples on one surface are compared with their nearest points anywhere on the opposite surface.
Our mesh-distance convention adapts point-set evaluation~\cite{pointsetgen2016} to continuous surfaces.

For mesh CD, HD, and NC, let $Q\subset S$ and $Q^*\subset S^*$ contain 100,000 area-uniform samples each, with a fixed UID-dependent seed derived from 42.

\paragraph{Mesh CD.}
The bidirectional mean uses unsquared point-to-surface distances:
\begin{equation}
 \mathrm{CD}=\frac12\left[
 \frac1{|Q|}\sum_{q\in Q}d(q,S^*)+
 \frac1{|Q^*|}\sum_{q\in Q^*}d(q,S)\right].
\end{equation}
\paragraph{Mesh HD.}
The sampled bidirectional maximum is
\begin{equation}
 \mathrm{HD}=\max\!\left\{\max_{q\in Q}d(q,S^*),\;
                         \max_{q\in Q^*}d(q,S)\right\}.
\end{equation}
Because the maximum is taken over sampled query points, HD estimates the worst surface discrepancy at the chosen sampling density.

\paragraph{Normal consistency.}
Write $\Pi_S(x)$ for a nearest point on $S$ and $n_S(x)$ for its unit face normal.
The orientation-insensitive score is
\begin{equation}
 \begin{aligned}
 \mathrm{NC}=\frac12\bigg[&\frac1{|Q|}\sum_{q\in Q}
 |n_S(q)^\top n_{S^*}(\Pi_{S^*}(q))|\\
 &+\frac1{|Q^*|}\sum_{q\in Q^*}
 |n_{S^*}(q)^\top n_S(\Pi_S(q))|\bigg].
 \end{aligned}
\end{equation}
The absolute dot products ignore normal orientation; larger values indicate better agreement between corresponding surface normals.

\paragraph{Vertex-to-surface distance.}
Vertex fidelity is evaluated directly on the generated vertices:
\begin{equation}
 \mathrm{V2S}=\frac1{|V|}\sum_{v\in V}d(v,S^*).
\end{equation}

\paragraph{Density-aware Chamfer distance.}
For density-aware Chamfer distance~\cite{dcd2021}, let $\nu(v)$ be the nearest reference vertex and $n(\nu(v))$ the number of generated vertices assigned to it.
The directional term is
\begin{equation}
 D(V,V^*)=\frac1{|V|}\sum_{v\in V}
 \left[1-\frac{\exp(-1000\|v-\nu(v)\|_2^2)}{n(\nu(v))}\right].
\end{equation}
The symmetric vertex-layout score is
\begin{equation}
 \mathrm{DCD}=\frac12\left[D(V,V^*)+D(V^*,V)\right].
\end{equation}
We use unique vertices, frequency exponent one, and no point-number ratio weights.
The reverse term exchanges the generated and reference vertex sets and recomputes nearest-vertex assignments and multiplicities.
The distance factor penalizes displaced vertices, while the multiplicity factor penalizes several vertices concentrating around the same nearest vertex.

\paragraph{Surface F-score.}
Surface F-score measures surface agreement after the generated vertices are connected into a mesh.
It complements V2S and DCD by testing whether the vertex layout supports a geometrically faithful generated surface, although the result also depends on the connectivity model.
Following precision--recall evaluation~\cite{fscore2019}, 20,000 area-uniform samples per surface estimate the fraction within $\tau=1/128$ of the opposite continuous surface in each direction.
The octree spans $[-1,1]^3$ at resolution $1024^3$, so its finest-level voxel width is $\Delta=2/1024$ and the matching threshold is $\tau=4\Delta$.
Denoting these sample sets by $U\subset S$ and $U^*\subset S^*$, define
\begin{equation}
 \begin{aligned}
 P_\tau&=\frac1{|U|}\sum_{u\in U}\mathbf1[d(u,S^*)\le\tau],\\
 R_\tau&=\frac1{|U^*|}\sum_{u\in U^*}\mathbf1[d(u,S)\le\tau].
 \end{aligned}
\end{equation}
Here $P_\tau$ measures the fraction of generated-surface samples close to the reference, and $R_\tau$ measures reference-surface coverage.
Their harmonic mean is
\begin{equation}
 F_\tau=\begin{cases}
 \dfrac{2P_\tau R_\tau}{P_\tau+R_\tau}, & P_\tau+R_\tau>0,\\
 0, & P_\tau+R_\tau=0.
 \end{cases}
\end{equation}
Tables report $100F_\tau$ as a percentage.

\subsection{DenseMesh Quantitative Results}
\label{sec:dense_results}

\begin{table}[!t]
\centering
\small
\caption{\textbf{Final mesh quality on DenseMesh.} $\dagger$Controlled reimplementation; $\ddagger$image + sparse-voxel input.}
\label{tab:dense_mesh}
\setlength{\tabcolsep}{5pt}
\renewcommand{\arraystretch}{1.06}
\begin{tabular}{@{}lrrr@{}}
\toprule
Method & CD$\downarrow$ & HD$\downarrow$ & NC$\uparrow$ \\
\midrule
\multicolumn{4}{l}{\textit{Autoregressive}} \\
EdgeRunner~\cite{edgerunner2024} & 24.23 & 31.48 & 0.790 \\
TreeMeshGPT~\cite{treemeshgpt2025} & 43.78 & 40.01 & 0.765 \\
BPT~\cite{bpt2024} & 58.81 & 47.97 & 0.801 \\
DeepMesh~\cite{deepmesh2025} & 43.98 & 40.83 & 0.809 \\
FastMesh~\cite{fastmesh2025} & 45.37 & 26.28 & 0.711 \\
MeshRipple~\cite{meshripple2025} & 34.46 & 38.18 & 0.758 \\
\midrule
\multicolumn{4}{l}{\textit{Diffusion}} \\
MeshFlow~\cite{meshflowmeta2026} & 11.93 & 8.42 & 0.800 \\
LATO.2~\cite{lato22026}$^{\ddagger}$ & 6.34 & 9.65 & 0.824 \\
Nexus (16K)$^{\dagger}$ & 4.14 & 10.91 & 0.888 \\
Nexus (40K)$^{\dagger}$ & 3.78 & 10.38 & 0.893 \\
\rowcolor{papergray} TOPFlow (16K) & \underline{3.05} & \underline{8.40} & \underline{0.921} \\
\rowcolor{papergray} TOPFlow (40K) & \textbf{2.64} & \textbf{7.15} & \textbf{0.925} \\
\bottomrule
\end{tabular}
\end{table}

\begin{table}[!t]
\centering
\small
\caption{\textbf{Vertex generation quality on DenseMesh.} $\dagger$Controlled reimplementation.}
\label{tab:dense_vertex}
\setlength{\tabcolsep}{5pt}
\renewcommand{\arraystretch}{1.06}
\begin{tabular}{@{}lrrr@{}}
\toprule
Method & V2S$\downarrow$ & DCD$\downarrow$ & F$\uparrow$ \\
\midrule
Nexus (16K)$^{\dagger}$ & 2.76 & 0.517 & 89.02 \\
Nexus (40K)$^{\dagger}$ & 2.67 & 0.517 & 90.43 \\
\rowcolor{papergray} TOPFlow (16K) & \underline{2.38} & \underline{0.511} & \underline{93.44} \\
\rowcolor{papergray} TOPFlow (40K) & \textbf{2.34} & \textbf{0.511} & \textbf{94.30} \\
\bottomrule
\end{tabular}
\end{table}

\cref{tab:dense_mesh,tab:dense_vertex} report final-mesh quality and vertex quality on DenseMesh, using a 4,000 vertex-count condition for Nexus and TOPFlow.
The mesh results in \cref{tab:dense_mesh} show that TOPFlow (40K) reduces CD and HD over Nexus (40K) by 30.1\% and 31.1\%, respectively, and raises NC from 0.893 to 0.925.
The vertex results in \cref{tab:dense_vertex} complement this comparison: TOPFlow improves V2S, DCD, and surface F-score over Nexus at both input densities.

\subsection{Temperature Report and Uncertainty}
\label{sec:temperature_report}

\paragraph{Purpose and evaluation protocol.}
Main \cref{fig:temperature} finds favorable vertex layouts near $(T_K,T_C,T_F)=(0,0,0)$.
We test whether this trend generalizes to unseen objects and whether separate factor tuning improves on a shared temperature.
The 128 grid-search objects serve as the \emph{development set} for selecting settings by mean vertex DCD.
A disjoint \emph{report set} of 256 objects evaluates the fixed selections, avoiding evaluation on the objects used for tuning.
Each set is equally divided among four reference-vertex-count complexity groups to cover different geometric complexities.
All comparisons use fixed weights and 40K inputs.

\paragraph{Joint versus shared-temperature selection.}
To assess the benefit of separate factor tuning, we select a joint optimum $J$ over all 125 triples in $\{0,0.25,0.5,0.75,1\}^3$ and a shared-temperature optimum $U$ over the five triples $(T,T,T)$ in that grid.
Development DCD selects $J=(0,0.25,0)$ and $U=(0,0,0)$, with means 0.2610 and 0.2635, respectively.
These choices are fixed before report evaluation.
Below, $000$ and $111$ denote $(0,0,0)$ and $(1,1,1)$.
\cref{tab:temperature_report} compares $J$ and all eight triples in $\{0,1\}^3$ on the report set.
These triples test zero versus unit temperature for each factor with the others fixed; vertex and mesh metrics show how the choices affect both generation stages.

\begin{table}[t]
\centering\small
\caption{\textbf{Temperature comparison on Toys4K.} The upper and lower blocks report vertex and mesh quality, respectively.}
\label{tab:temperature_report}
\setlength{\tabcolsep}{5pt}
\renewcommand{\arraystretch}{0.95}
\begin{tabular}{@{}lrrr@{}}
\toprule
$(T_K,T_C,T_F)$ & V2S$\downarrow$ & DCD$\downarrow$ & F$\uparrow$ \\
\midrule
\rowcolor{papergray} (0,0,0) & \textbf{1.41} & \underline{0.2742} & 97.02 \\
(0,0,1) & 1.50 & 0.2827 & 96.88 \\
(0,0.25,0) & \underline{1.41} & \textbf{0.2739} & \textbf{97.37} \\
(0,1,0) & 1.44 & 0.2820 & \underline{97.14} \\
(0,1,1) & 1.55 & 0.2909 & 97.08 \\
(1,0,0) & 1.51 & 0.3041 & 96.81 \\
(1,0,1) & 1.57 & 0.3212 & 96.40 \\
(1,1,0) & 1.56 & 0.3134 & 96.76 \\
(1,1,1) & 1.66 & 0.3327 & 96.31 \\
\midrule
$(T_K,T_C,T_F)$ & CD$\downarrow$ & HD$\downarrow$ & NC$\uparrow$ \\
\midrule
\rowcolor{papergray} (0,0,0) & 3.01 & 6.49 & 0.973 \\
(0,0,1) & 3.09 & 6.63 & 0.972 \\
(0,0.25,0) & \textbf{2.29} & \textbf{5.35} & \textbf{0.974} \\
(0,1,0) & 2.51 & 6.27 & \underline{0.974} \\
(0,1,1) & \underline{2.50} & \underline{5.91} & 0.973 \\
(1,0,0) & 2.72 & 6.12 & 0.973 \\
(1,0,1) & 3.56 & 7.16 & 0.971 \\
(1,1,0) & 3.10 & 6.44 & 0.972 \\
(1,1,1) & 3.11 & 6.82 & 0.971 \\
\bottomrule
\end{tabular}
\end{table}

\paragraph{Independent report results.}
On the report set, $000$ reduces mean DCD from 0.3327 at $111$ to 0.2742, a 17.6\% reduction.
We assess this difference using a stratified paired bootstrap with 10,000 resamples: objects are resampled within each complexity group, with all compared settings kept paired for each object.
The mean difference $000-111$ is $-0.05853$, with a 95\% confidence interval of $[-0.06577,-0.05118]$.
The interval is entirely below zero, supporting the low-temperature trend on unseen report objects.
By comparison, $J$ yields DCD 0.2739 versus 0.2742 for $U=000$, an improvement of only 0.10\%.
The paired difference $J-000$ is $-0.000274$, with a 95\% interval of $[-0.002533,0.002061]$.
This interval includes zero, so joint tuning's development-set advantage does not establish a clear report-set DCD advantage.

\paragraph{Interpretation and default setting.}
The report results support $000$ as a simple vertex-generation default: it lowers DCD relative to $111$, with no clear DCD disadvantage relative to $J$.
However, $J$ has lower mean mesh CD and HD and higher mean surface F-score in \cref{tab:temperature_report}.
\cref{sec:temperature_proofs} provides the theoretical mechanism: zero temperature retains conditional maximizers of count, allocation, and layout, reducing averaging among competing patterns in the posterior mean.
Unique maxima yield one pattern, while ties retain multiple patterns.
These properties explain the flow change but guarantee neither full-pattern MAP nor minimum geometric error; the default is supported empirically by the independent vertex-layout comparison.

\section{Additional Qualitative Comparisons}
\label{sec:additional_qualitative}

\subsection{Robustness to Input Point Count}
\label{sec:surface_point_robustness}

\begin{figure}[!ht]
\centering
\includegraphics[width=0.90\textwidth]{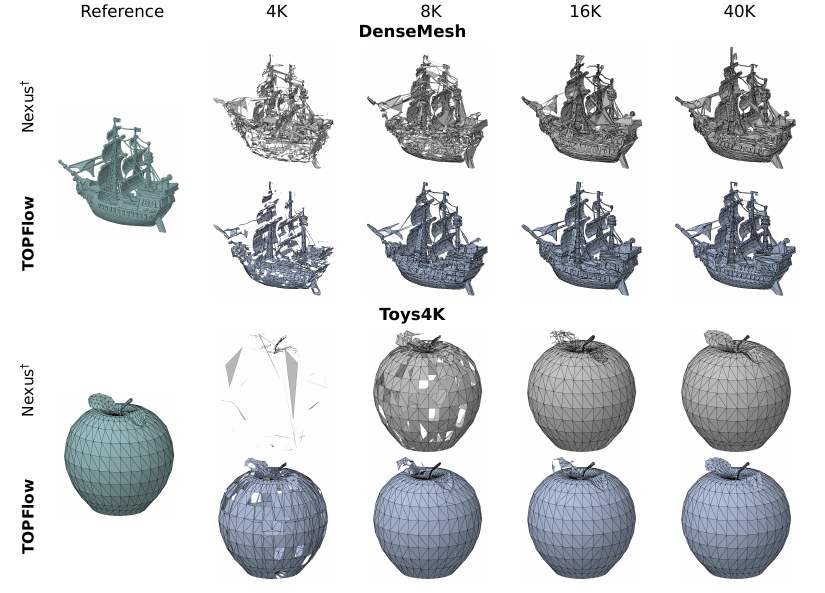}
\caption{\textbf{Robustness to input point count.} Nexus and TOPFlow are trained with 40K surface points and evaluated at 4K, 8K, 16K, and 40K. $^{\dagger}$Controlled reimplementation.}
\label{fig:surface_point_robustness}
\end{figure}

\cref{fig:surface_point_robustness} examines how reducing the number of conditioning points affects mesh generation.
Both models are trained with 40K surface points with normals; 4K, 8K, and 16K inputs are used only at inference.
The DenseMesh ship tests retopology from a high-poly surface whose geometry and normals differ from the low-poly training distribution, while the Toys4K apple provides an example with complexity comparable to the training data.

At 4K, TOPFlow retains more of the ship's hull and the apple's surface, whereas Nexus produces a fragmented hull and largely disconnected apple geometry.
At 8K, TOPFlow yields a more continuous apple surface and more coherent ship structure.
Denser inputs improve both methods, but TOPFlow preserves comparatively complete surfaces and coherent local edge arrangements across the two examples.
The comparison therefore indicates robustness to both unseen input point counts and high-poly surface geometry.

Sparse inputs still lose fine structures and leave local holes.
Training with mixed input point counts could improve performance with sparse conditioning point clouds.

\subsection{Level-of-Detail Control}
\label{sec:level_of_detail_control}

\begin{figure}[!ht]
\centering
\includegraphics[width=0.90\textwidth]{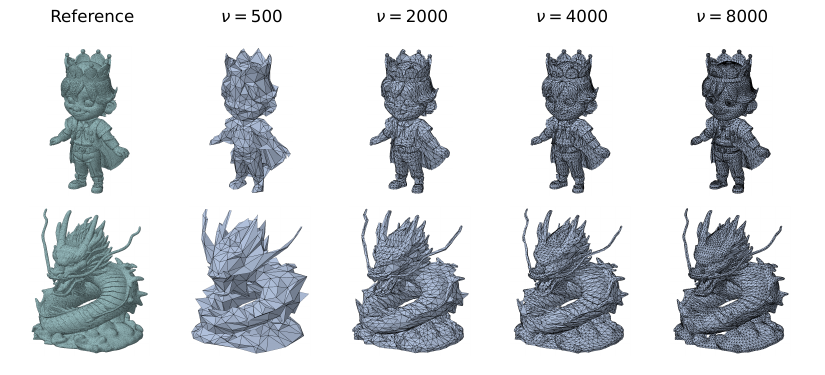}
\caption{\textbf{Level-of-detail control.} Meshes generated with vertex-count conditions $\nu\in\{500,2000,4000,8000\}$.}
\label{fig:level_of_detail_control}
\end{figure}

The vertex-count condition $\nu$ allows TOPFlow to generate meshes at different levels of detail from one conditioning point cloud.
\cref{fig:level_of_detail_control} illustrates this control on a DenseMesh character and a coiled dragon.
At smaller $\nu$, compact meshes preserve the overall silhouettes and major components.
Increasing $\nu$ refines the character's facial features and clothing, as well as the dragon's head, horns, and coiled body.

These examples show that the same model supports both compact and detailed outputs for high-poly retopology.
As a soft condition, $\nu$ guides output complexity while the final vertex count is determined by the generated occupancy.

\subsection{Further Comparisons across Datasets}
\label{sec:further_qualitative_comparisons}

\begin{figure}[!ht]
\centering
\includegraphics[width=\textwidth]{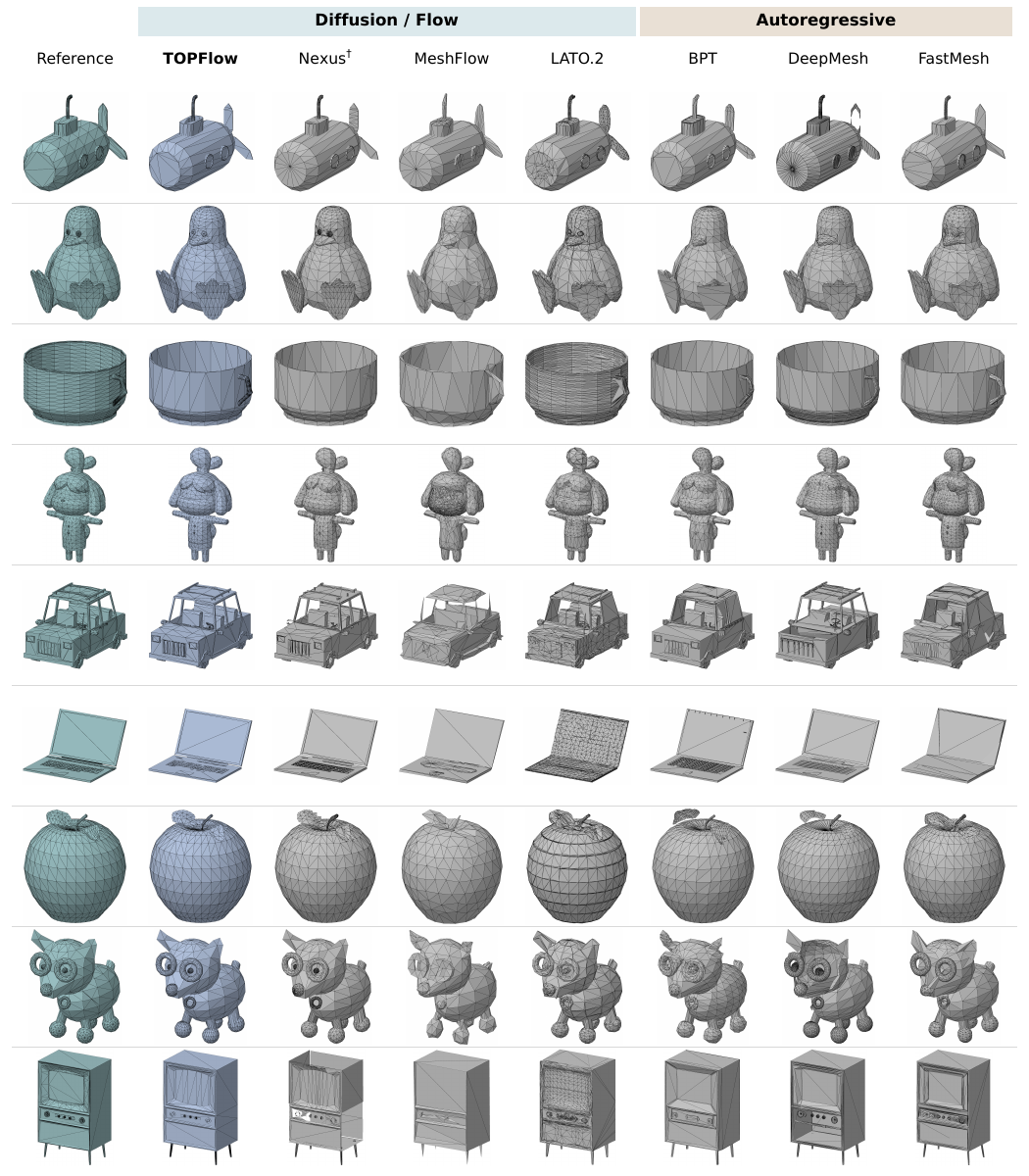}
\caption{\textbf{Additional mesh comparisons on Toys4K.} Nexus and TOPFlow use 40K inputs. $^{\dagger}$Controlled reimplementation.}
\label{fig:supp_toys4k}
\end{figure}

\begin{figure}[!ht]
\centering
\includegraphics[width=\textwidth]{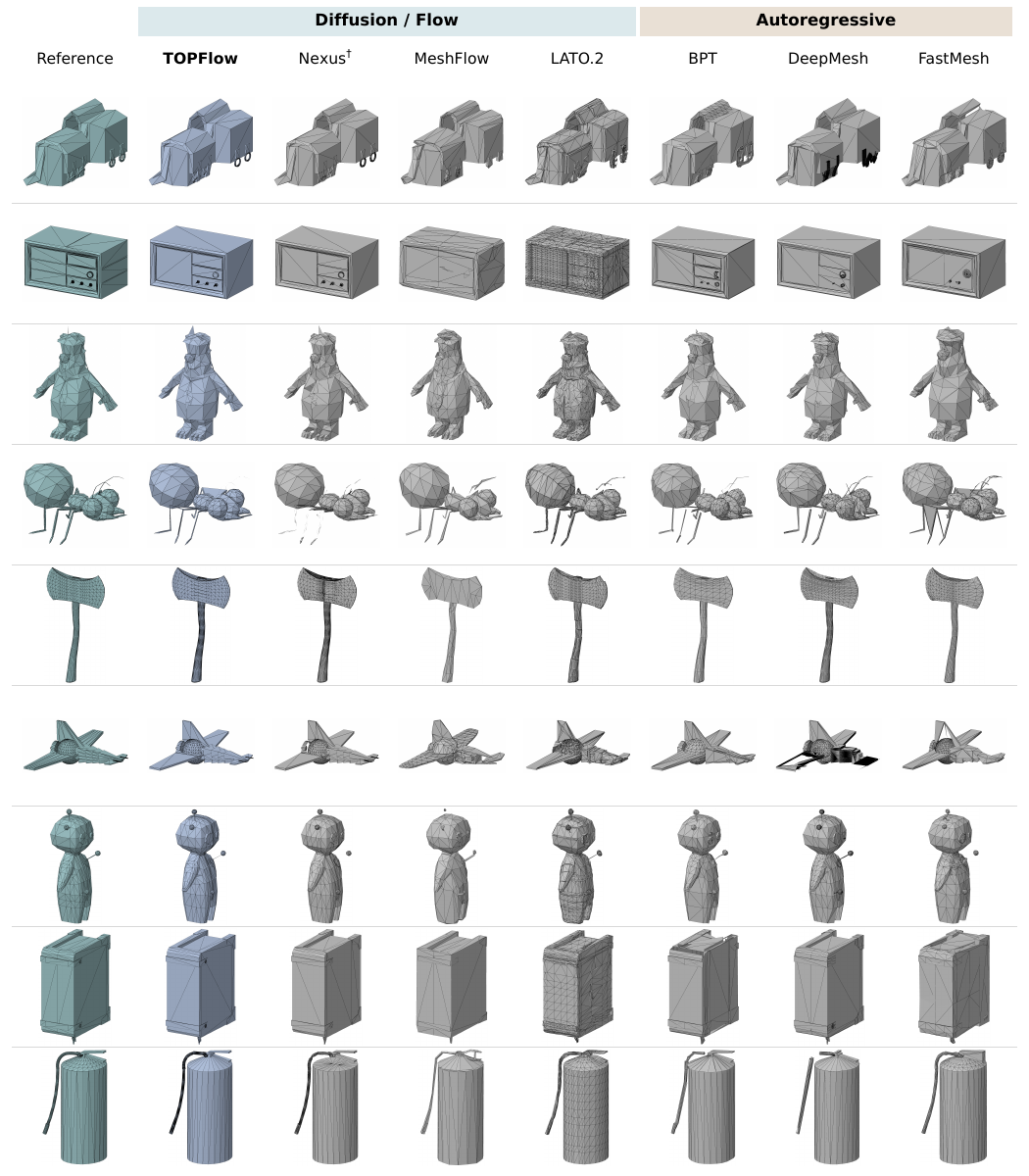}
\caption{\textbf{Additional mesh comparisons on Objaverse.} Nexus and TOPFlow use 40K inputs. $^{\dagger}$Controlled reimplementation.}
\label{fig:supp_objaverse}
\end{figure}

\begin{figure}[!ht]
\centering
\includegraphics[width=\textwidth]{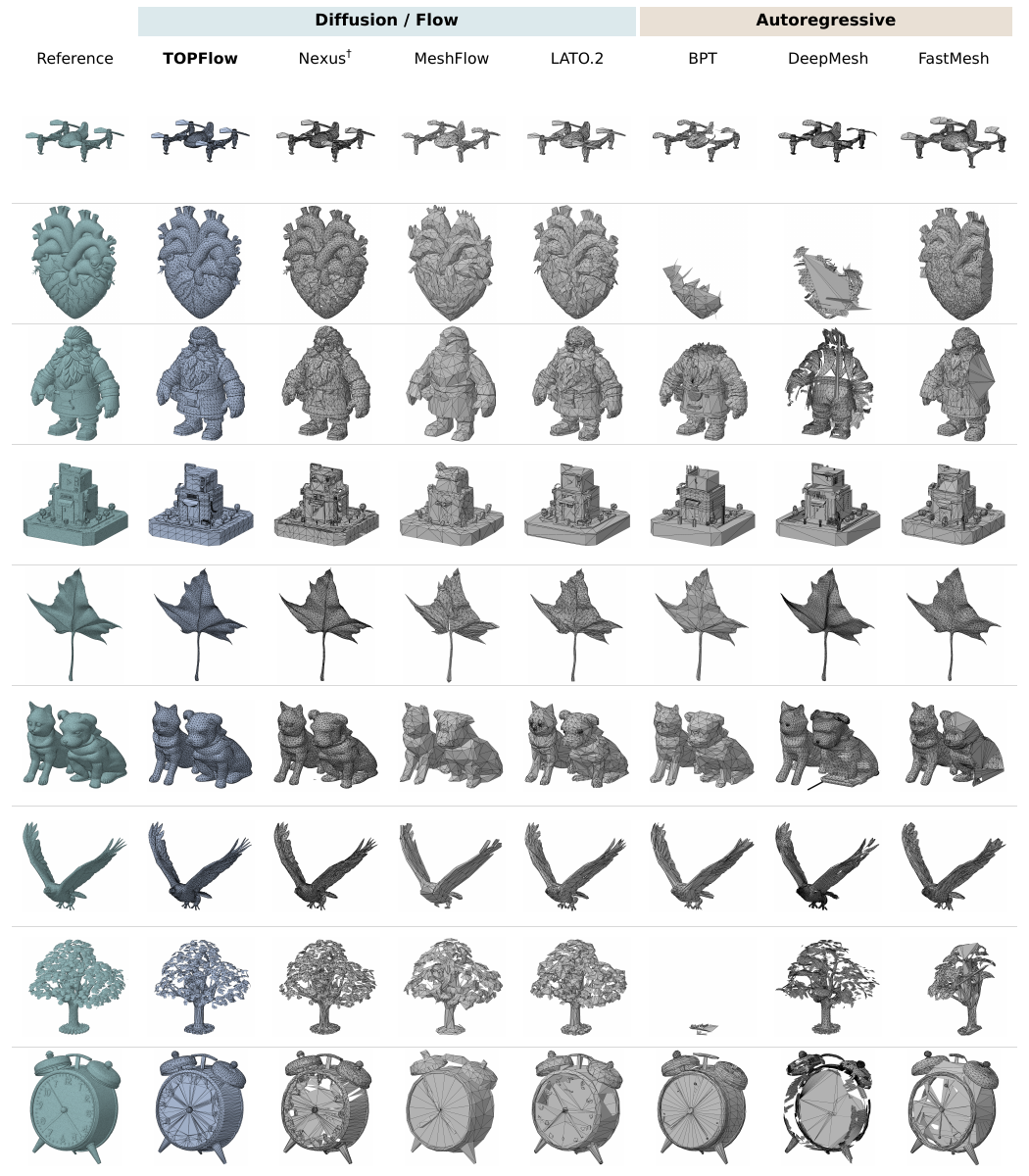}
\caption{\textbf{Additional mesh comparisons on DenseMesh.} Nexus and TOPFlow use 40K inputs. $^{\dagger}$Controlled reimplementation.}
\label{fig:supp_densemesh}
\end{figure}

\cref{fig:supp_toys4k,fig:supp_objaverse,fig:supp_densemesh} extend the main qualitative comparison with several additional objects per dataset, covering surface completeness, small structures, and local edge arrangements.
On Toys4K, TOPFlow preserves the car frame, laptop panels, and apple silhouette, while some baselines distort the car or introduce irregular connections.
The cabinet further illustrates the preservation of thin supports and planar panels.

On Objaverse, TOPFlow retains the ant's body and legs, the axe's narrow handle, and the appliance contours in \cref{fig:supp_objaverse}.
Several baselines capture the broad shapes but simplify small components or fragment slender structures, highlighting the importance of preserving detail alongside the overall silhouette.

On DenseMesh, TOPFlow produces compact meshes while preserving the heart's overall structure, the castle silhouette, and the shapes of the leaves and paired animals in \cref{fig:supp_densemesh}.
It also retains a more continuous clock body than several baselines.
These comparisons highlight TOPFlow's ability to preserve overall shapes and fine structures with coherent local edge arrangements across diverse objects.

\newpage
\onecolumn
\clearpage
\clearpage
\clearpage
\clearpage
\section{Evaluation Dataset Overviews}
\label{sec:dataset_overviews}

\begin{figure}[!ht]
\centering
\includegraphics[width=\textwidth]{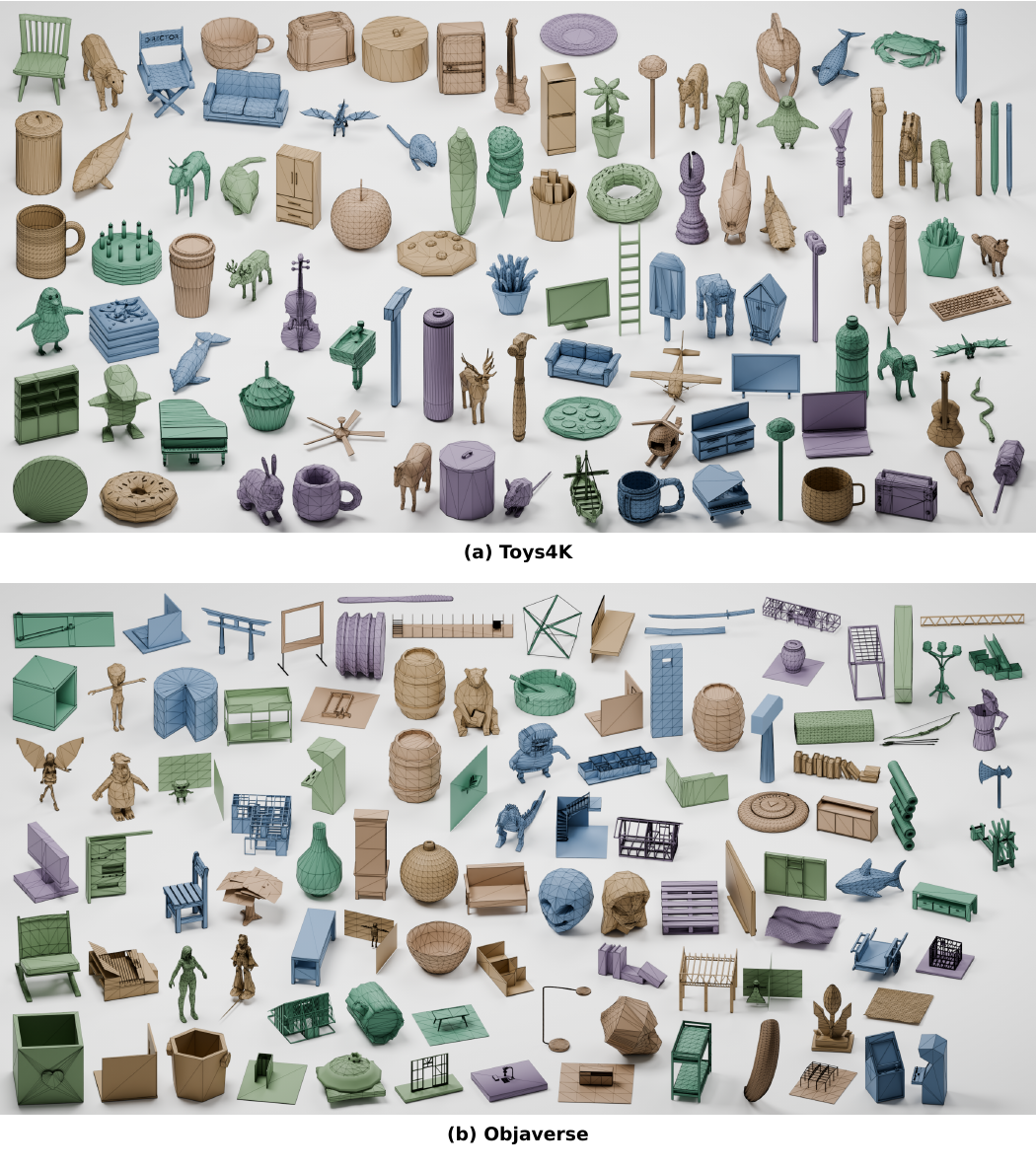}
\caption{\textbf{Random mesh examples from the evaluation datasets.} Each panel shows 100 reference meshes sampled from (a) Toys4K and (b) Objaverse, with display colors and visible polygon edges.}
\label{fig:dataset_overviews}
\end{figure}
\clearpage
\twocolumn

\FloatBarrier

\end{document}